\documentclass[10pt,twocolumn,letterpaper]{article}

\usepackage[pagenumbers]{cvpr} 

\usepackage{graphicx}
\usepackage{amsmath}
\usepackage{amssymb}
\usepackage{booktabs}
\usepackage{tabularx}
\usepackage{balance}
\usepackage{multirow}
\usepackage{colortbl}
\usepackage{arydshln}
\usepackage{float}
\usepackage[dvipsnames]{xcolor}
\usepackage[T1]{fontenc}
\usepackage{ulem}

\definecolor{lightred}{RGB}{255, 230, 230}
\definecolor{lightblue}{RGB}{230, 240, 255}

\usepackage[pagebackref,breaklinks,colorlinks,linkcolor=red,urlcolor=blue,citecolor=teal]{hyperref}

\definecolor{verylightgray}{HTML}{E0E0E0}

\newcolumntype{a}{>{\columncolor{verylightgray}}c}

\usepackage[capitalize]{cleveref}
\crefname{section}{Sec.}{Secs.}
\Crefname{section}{Section}{Sections}
\Crefname{table}{Table}{Tables}
\crefname{table}{Tab.}{Tabs.}

\def\cvprPaperID{39} 
\def\confName{CVPR}
\def\confYear{2025}

\begin{document}




\title{Towards Application-Specific Evaluation of Vision Models: Case Studies in Ecology and Biology}









\author{Alex Hoi Hang Chan$^{*1,2,3}$, Otto Brookes$^{*4,5}$, Urs Waldmann$^{1,6}$, Hemal Naik$^{1,7}$, Iain D. Couzin$^{1,2,3}$\\ Majid Mirmehdi$^{4}$, Noël Adiko Houa$^{5}$, Emmanuelle Normand$^{5}$, Christophe Boesch$^{5}$, Lukas Boesch$^{5}$\\ Mimi Arandjelovic$^{9}$, Hjalmar Kühl$^{8}$, Tilo Burghardt$^{\dagger4}$, Fumihiro Kano$^{\dagger1,2,3}$\\\\
\normalsize $^1$Centre for the Advanced Study of Collective Behaviour,  University of Konstanz, \\
\normalsize $^2$Department of Collective Behavior, Max Planck Institute of Animal Behavior,\\
\normalsize $^3$Department of Biology, University of Konstanz, 
\normalsize $^4$School of Computer Science, University of Bristol, \\
\normalsize $^5$Wild Chimpanzee Foundation, 
\normalsize $^6$Department of Computer and Information Science, University of Konstanz, \\
\normalsize $^7$Department of Ecology of Animal Societies, Max Planck Institute of Animal Behavior, \\
\normalsize $^8$Senckenberg Museum of Natural History Goerlitz,
\normalsize $^9$Max Planck Institute for Evolutionary Anthropology
\\ \small $^{*,\dag}$contributed equally;
\small hoi-hang.chan@uni-konstanz.de, otto.brookes@bristol.ac.uk
}

\maketitle

\begin{abstract}\vspace{-8pt}
Computer vision methods have demonstrated considerable potential to streamline ecological and biological workflows, with a growing number of datasets and models becoming available to the research community. However, these resources focus predominantly on evaluation using machine learning metrics, with relatively little emphasis on how their application impacts downstream analysis. We argue that models should be evaluated using application-specific metrics
that directly represent model performance in the context of its final use case. To support this argument, we present two disparate case studies: (1) estimating chimpanzee abundance and density with camera trap distance sampling when using a video-based behaviour classifier and (2) estimating head rotation in pigeons using a 3D posture estimator. We show that even models with strong machine learning performance (e.g., 87.82\% mAP) can yield data that leads to substantial discrepancies in abundance estimates compared to expert-derived data. Similarly, the highest-performing models for posture estimation do not  produce the most accurate inferences of gaze direction in pigeons. Motivated by these findings, we call for researchers to integrate application-specific metrics in ecological/biological datasets, allowing for models to be benchmarked in the context of their downstream application and to facilitate better integration of models into application workflows.








\end{abstract}

\vspace{-18pt}\section{Introduction}\label{sec:introduction}\vspace{-6pt}


Computer vision (CV) has emerged as a powerful tool in ecology and biology, with the potential to dramatically reduce the effort required for data extraction~\cite{tuia2022perspectives}. 
These advantages have important implications for many disciplines, including biodiversity monitoring~\cite{pollock2025harnessing}, conservation~\cite{tuia2022perspectives}, behavioural ecology~\cite{couzin2022emerging}, neurobiology~\cite{mathis2020deep} and more.

In recent years, there has been significant development of datasets and algorithms~\cite{ng2022animal, chen2023mammalnet, liu2023lote, 3D-POP, 3dmuppet, brookes2023triple, kholiavchenko2024deep, brookes2024chimpvlm, naik2024bucktales, johanns2022automated, haucke2022overcoming,DLC,SLEAP} for animal studies. While these algorithms are rigorously evaluated using machine learning (ML) metrics (e.g., accuracy, mean average precision (mAP), root mean squared error (RMSE)), their impact on ecological and biological research remains largely underexplored. Only a small number of studies have examined how CV solutions translate to real-world applications and their effectiveness in deriving meaningful measurements in downstream analysis~\cite {whytock2021robust,pantazis2024deeplearning,christensen2024moving, henrich2024semi, pollock2025harnessing,naik2024bucktales,chan2024comparison}. 
However, the emergence of application-driven ML learning~\cite{rolnick_position_2024} and its adoption by major conferences~\cite{ICML,ICCV,CVPR} is beginning to bridge this gap and it is a timely opportunity to evaluate the practical usability of ML models in their intended applications.



In this paper, we argue models should be evaluated using application-specific metrics that reflect effectiveness in their intended ecological or biological use cases, in addition to standard ML metrics. To support this argument, we present two case studies with both methods of evaluation. First, we evaluate a behaviour classification model for removing video segments containing behaviours known to bias population abundance estimates calculated using camera trap distance sampling (CTDS). We demonstrate that; (a) high-performing ML models (e.g., 87.82\% mAP) can output data that, when used in CTDS, produce abundance estimates that differ notably from those obtained through manual expert annotation and; (b) the way model predictions interact with statistical approaches, such as CTDS, is difficult to predict.
Second, we employ 3D posture estimation algorithms to infer attention in pigeons by estimating head rotation. We find that; (c) there is a mismatch between the accuracy of keypoint estimation with rotation estimation, and; (d) mismatches can result in misleading conclusions about which model is the most appropriate for the final application.


To better align models with their intended use case, we recommend that application-specific metrics are included in existing/future datasets and reported on when proposing new models, as complementary metrics to existing ML benchmarks. Building on recent calls for closer interdisciplinary collaboration~\cite{tuia2022perspectives, pollock2025harnessing}, we highlight application-specific metrics as a practical and accessible starting point for joint efforts between ML and ecology/biology researchers that stands to benefit both fields. 







\section{Related Work}
As early as 2012, Wagstaff~\cite{wagstaff2012machine} highlighted over-reliance on benchmark datasets, lack of method Interpretability, and limited capacity for real-world application in the field of machine learning. More recent commentaries have highlighted limitations with the current benchmarking system~\cite{liao2021we}, and need for alternative evaluation procedures~\cite{raji2021ai}. Recently, Rolnick et al.~\cite{rolnick_position_2024} formulated a paradigm to separate ML research into methods-driven ML and application-driven ML. The latter is intended to overcome the aforementioned limitations by promoting closer collaboration with end users, conservative use of standardized datasets, and evaluation metrics tailored to specific domains. In biological and ecological studies, there is increasing interest in evaluating ML models in an application-specific context~\cite {whytock2021robust,pantazis2024deeplearning,christensen2024moving, henrich2024semi, pollock2025harnessing,naik2024bucktales,chan2024comparison,hu2023labgym,kaneko2024deciphering}.







\textbf{Ecology}. Whytock et al.~\cite{whytock2021robust} trained the ResNet50 architecture to classify 26 Central African mammal and bird species with a top-1 accuracy of 77.63\% and showed that predictions were equivalent to expert labelling when considering species richness, occupancy and activity patterns. Pantazis et al.~\cite{pantazis2024deep} assess the impact of model architecture, label noise, and dataset size on the same ecological metrics, showing that model architecture is of minimal importance although noise and dataset size is significant. Henrich et al.~\cite{henrich2024semi} perform semi-automatic extraction of distances between animal and camera with a deep learning model and predict population densities for 10 species using CTDS. They show densities predicted with manual and semi-automatic processes do not differ significantly.


\textbf{Biology}. 
Chistensen et al.~\cite{christensen2024moving} highlight that ML metrics (F1-score) of a behavioural classifier from accelerometer data can be misleading, and even low accuracy can lead to a reliable inference of behavioural states in wild birds. Benchmarking results from the BuckTales dataset~\cite{naik2024bucktales} suggest that commonly used MOT metric, MOTA, is not suitable to evaluate animal tracking algorithms, as it does not consider ID switches, which is highly relevant for biological analysis. Recent studies have also evaluated ML-based pipelines by their ability to predict simple biological hypotheses in animal behaviour~\cite{chan2024comparison,christensen2024moving}, or detect differences in medical treatment groups, e.g~\cite{hu2023labgym,kaneko2024deciphering}. Specifically for animal posture estimation, to the best of our knowledge, there is limited work evaluating the relationship between the accuracy of keypoint estimation models and downstream applications, despite the widespread application of the method.




\begin{figure}[!ht]
\centering
\includegraphics[width=1.0\linewidth]{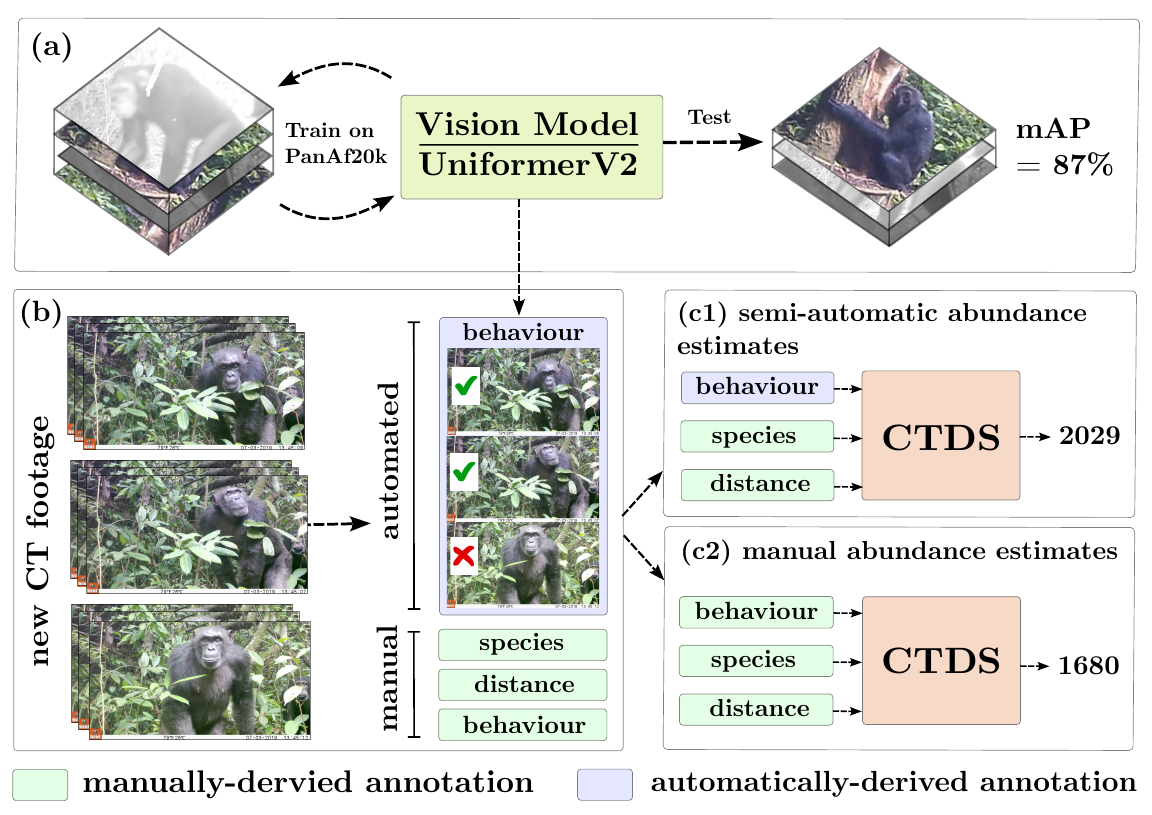}
\caption{\textmd{\textbf{{Case Study 1 Overview}:} \textbf{(a)} Train and test a strong camera reactivity classifier on the PanAf20k dataset. Note the use of an ML metric here; \textbf{(b)} Apply the trained model to classify clips containing camera reactivity from new camera trap video for removal/filtering; \textbf{(c)} Estimate abundance -- an application-specific metric -- with CTDS using model-derived behaviour annotations (c1) and compare with expertly-derived annotations (c2). The semi-automatic approach leads to an overestimation of abundance by 20.77\% when compared with manual annotation (see Tab.~\ref{tab:density_abundance_est} for full details).}}
\label{fig:overview}
\end{figure}

\begin{table*}[!h]
\footnotesize
\centering
\begin{tabular}{ccccllllll} \toprule
    & \multirow{2}{*}{\bf Detection Function} & \multirow{2}{*}{\bf Removal Method} & \multirow{2}{*}{\bf Camera Reaction} & \multicolumn{3}{c}{\bf Abundance} & \multicolumn{3}{c}{\bf Density (ind./km$^2$)} \\
    \cmidrule(l){5-7}\cmidrule(l){8-10}
  & & & & {\bf Estimate; SE} & \multicolumn{2}{c}{\bf 95\% CIs} & {\bf Estimate; SE} & \multicolumn{2}{c}{\bf 95\% CIs} \\ \midrule
1 & & None & Yes & 2575; 921 & 1122 & 4981 & 0.47; 0.16 & 0.24 & 0.80\\
\rowcolor{lightred}
2 & Hr1 & Manual & No & 1680; 755 & 578 & 3235 & 0.34; 0.14 & 0.13 & 0.64\\
\rowcolor{lightblue}
3 &  & Auto & No & 2029; 767 & 703 & 3658 & 0.36; 0.17 & 0.17 & 0.68\\
  \hdashline
4 & & None & Yes & 2529; 892 & 1313 & 4332 & 0.49; 0.18 & 0.21 & 0.83\\
\rowcolor{lightred}
5 & Hn1 & Manual & No & 1954; 943 & 693 & 4015 & 0.36; 0.16 & 0.12 & 0.71\\
\rowcolor{lightblue}
6 &  & Auto & No & 2235; 909 & 745 & 3806 & 0.38; 0.15 & 0.16 & 0.63\\
  \bottomrule
\end{tabular}
\caption{Density \& abundance estimations calculated with CTDS under two different detection functions (Hr1 and Hn1) with standard errors (SE) and 95\% confidence intervals (CIs) shown. It compares two methods for the removal of video clips showing camera reactivity: \colorbox{lightred}{Manual} (clips identified and removed by a human expert) and \colorbox{lightblue}{Auto} (clips automatically identified for removal by a behaviour classifier). None indicates no clips were removed. Failing to remove clips with camera reactivity leads to the overestimation of density and abundance (row 1 vs. 2 and 4 vs. 5). The automated method does not remove \textit{all} camera reactivity clips leading to overestimates in density and abundance when compared with manual removal (row 2 vs. 3 and 5 vs. 6).
}
\label{tab:density_abundance_est}
\end{table*}

\section{Case study 1: Abundance \& Density Estimation of Chimpanzees}

\textbf{Overview}. CTDS is a common approach to estimating the abundance of a species through camera trapping~\cite{howe2017distance, delisle2021next, houa2022animal, delisle2023reducing}. It measures distances from cameras to detected animals and models detection probability as a function of distance to estimate population abundance and density. At its core is a detection function which describes how the probability of detecting an animal decreases with increasing distance from the camera. The data required by CTDS is difficult to obtain, necessitating the annotation of species, animal-camera-distances, and behaviour in camera trap videos.

In this case study, we automate one stage of the data acquisition pipeline: behaviour recognition. Camera reactivity behaviour is known to bias abundance estimates by causing chimps to remain in (i.e., attraction) or out (i.e., avoidance) of the cameras view, which can lead to overestimation and underestimation of population abundances, respectively~\cite{houa2022animal, delisle2023reducing}. Identifying and removing video clips where camera reactivity is observed currently represents the most effective method to debias estimates~\cite{delisle2023reducing}. Below, we describe the training and evaluation of a behaviour recognition model, and evaluate its ability to identify clips in which chimpanzees exhibit camera reactivity and the effect of using it on abundance estimates. 


\subsection{Experimental Setup}\label{sec:setup}

\textbf{Model Implementation \& Training}. We train UniformerV2~\cite{li2022uniformerv2} on a modified version of the PanAf20k dataset~\cite{brookes2024panaf20k} to classify camera reactivity (i.e., presence or absence of reaction) of chimpanzees in camera trap videos. To achieve this, we convert the labels of the dataset into a binary format: videos annotated with camera reactivity are assigned a positive label, while those without are assigned a negative label. We follow the training protocol detailed in~\cite{brookes2024panaf20k}, except we use a class balanced focal loss~\cite{cui2019class} to mitigate the effect of class imbalance (a 70:30 class imbalance between non-reactive and reactive videos exists). Macro-averaged mean average precision (mAP) is used for model evaluation.

\textbf{Abundance Estimation \& Model Evaluation}. To evaluate impact on downstream abundance estimates a different dataset comprising 413 videos from the Taï National Park, Côte d'Ivoire, is used. Each video is manually annotated with all the information required to generate abundance estimates using CTDS~\cite{houa2022animal}. The calculations are performed following the approach outlined by Howe et al.~\cite{howe2017distance}. The trained model is applied over 2 second segments of all videos (i.e., the same time intervals used during manual annotation), and clips classified as containing camera reactivity are removed prior to CTDS. Note that inference is performed in an out-of-distribution (OOD) setting since the camera trap locations here are unseen during training.




\subsection{Results}
First, the performance of the behaviour recognition model is reported, as evaluated using typical ML metrics (mAP). Then, the abundance estimates produced using the model filtered video footage are compared with those calculated using expert annotation and removal.
Results are reported for two different detection functions: half-normal (Hn1) and hazard rate (Hr1). Hn1 assumes a smooth decline in detection, while Hr1 allows for a flat detection zone followed by a steep drop. 
Note that all other data required for distance sampling (i.e., intra-video individual identities, species, distance from camera to individual etc.) are produced expertly. Thus, only the effect of automated behaviour classification on CTDS is examined.


\textbf{Result 1: Behaviour Recognition Performance}. 
 The model achieves an mAP score of 87.82\%,   
although there is a noticeable discrepancy in class-wise performance. While it effectively detects the absence of camera reactivity with 95.31\% AP, performance on detecting presence is lower at 73.95\%.  This discrepancy highlights the significance of class imbalance \textit{even} when imbalances are not extreme.

\begin{figure*}[!h]
\centering
\includegraphics[width=1.0\linewidth]{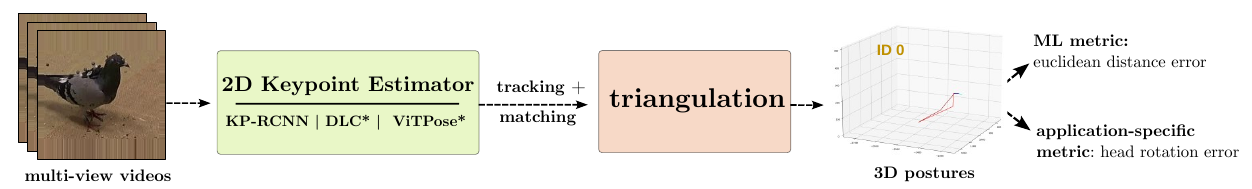}
\caption{\textmd{\textbf{{Case Study 2 Overview}. } We use the same pipeline proposed in 3D-MuPPET~\cite{3dmuppet}, by taking multi-view videos from 3D-POP~\cite{3D-POP}, then benchmark 3 different 2D posture detection architectures. We then run the same correspondence matching and triangulation algorithms to obtain 3D postures, which we use to compute ML metrics and application specific metrics. See original publication~\cite{3dmuppet} for detailed pipeline.}}
\label{fig:overview}
\end{figure*}




\textbf{Result 2: Abundance \& Density Estimation}. If camera reactivity clips are not removed prior to CTDS then population abundance and densities are \textit{significantly} overestimated (Tab.~\ref{tab:density_abundance_est}; 2575 vs. 1680 and 2529 vs. 1954 for Hr1 and Hn1, respectively). Across both detection functions, estimates are highest when no removal of camera reactivity clips is performed (row 1 and 4 vs. rest). Tab.~\ref{tab:density_abundance_est} also shows that using the behaviour classifier to remove camera reactivity clips still results in over estimation of abundance/density, across both detection functions. However, the degree of abundance overestimation is far greater in Hr1 (+20.77\%) than Hn1 (+14.38\%). Density is also overestimated but there is relatively little difference in the increase between detection functions. 
This difference shows that even when the \textit{same} model predictions are used to remove clips, the resulting abundance estimates can vary depending on the choice of detection function. This highlights that it is not just the number of correctly classified segments that matters — but \textit{which} segments are correctly identified and removed.







\section{Case study 2: Gaze Estimation in Pigeons}

\textbf{Overview.} 
Vision is a primary sensory channel for many species, and the direction of attention—where an animal directs its eyes or head (gaze)—provides crucial insights into how animals acquire information about their environments.
In this case study, we evaluate markerless 3D posture estimation methods by comparing their absolute Euclidean position errors with angular head orientation errors, the latter being a more meaningful measure for assessing the accuracy of inferred gaze direction.
Similar to humans, many bird species have foveas—specialized regions in the eye where visual acuity is highest~\cite{bringmann2019structure}. Recent studies have shown that by measuring a bird’s 3D head rotation, researchers can infer its gaze direction using both marker-based\cite{kano2022head, itahara2022corvid, nagy2023smart, delacoux2024fine} and markerless motion capture systems\cite{3D-POP, chimento2024peering}. When combined with simple behavioural experiments that estimate the location and extent of the bird’s visual field during object-directed attention, head orientation can serve as a reliable proxy for gaze direction~\cite{kano2022head,kano2018head,butler2018birds}.


\subsection{Experimental setup}
\textbf{Dataset \& Evaluation Metrics.} Here, we use the 3D-POP dataset~\cite{3D-POP}, a large scale 2D to 3D posture dataset in pigeon flocks, to reproduce benchmarks introduced in the 3D-MuPPET framework~\cite{3dmuppet} for 3D posture estimation. In 3D-MuPPET, the authors provide benchmark results. We refer to original publication for implementation details~\cite{3dmuppet}. in terms of RMSE and percentage correct keypoints (PCK). RMSE is calculated by taking the root mean squared of the Euclidean distances between all predicted and ground truth keypoints, while PCK is the percentage of predicted keypoints that fall within 5\% or 10\% of the maximum length of any two ground truth 3D keypoints\cite{3dmuppet}. In the current study, we only report these metrics for four head keypoints - not the nine keypoints reported in~\cite{3dmuppet}, so the comparison is fair with the alternative metric we propose below.


We propose an alternative metric that is more aligned to the biological use case, simply the absolute rotational error of the head. We employ the same method shown in~\cite{chimento2024peering} to calculate the rotation error of the head of each pigeon in each frame, in terms of the yaw, pitch and roll axis relative to the forward axis (axis from the mid-point between the eyes towards the beak).
The acceptable rotational error in head orientation depends on the spatial arrangement of gaze targets, such as conspecifics or objects of interest (e.g. food) within the observation arena. In most experimental setups, these targets are typically separated by more than $10^{\circ}$. Moreover, pigeons generally move their eyes within a narrow range—typically no more than $5^{\circ}$—when attending to distant objects~\cite{wohlschlager1993head,kano2022head}. Therefore, a head orientation error within $5^{\circ}$ from 3D posture tracking is considered an acceptable margin for reliably estimating gaze direction.
\subsection{Results}
We refer to Table \ref{tab:method_comparison} for the comparison between ML and application-specific metrics. Firstly, we find that based on the median head rotation error, all of the models compared are below $5^{\circ}$, confirming that the 3D-MuPPET framework is appropriate for the gaze estimation task. In this context, any of the architectures would have been within the acceptable accuracy, so the end user can consider other factors like inference speed or trajectory tracking accuracy when deciding which architecture to use (see~\cite{3dmuppet}). 

\begin{table*}[!h]
    \centering
    \resizebox{\textwidth}{!}{
    \begin{tabular}{lcccc|ccccc}
    \hline
    \multicolumn{5}{c|}{\textbf{ML Metrics}} & \multicolumn{5}{c}{\textbf{Application Specific Metrics }} \\

    \hline
    Method & RMSE (mm) & Median (mm) & PCK05 (\%) & PCK10 (\%) & RMSE Angles ($^\circ$) & Median Angles ($^\circ$) & Median Yaw ($^\circ$) & Median Pitch ($^\circ$) & Median Roll ($^\circ$) \\
    \hline
    3D-KP-RCNN   & 22.8  & 5.46  & 79.9 & 91.7 & 18.2  & 4.23  & 4.22  & 3.42  & 5.60  \\
    3D-DLC*      & 21.1  & 5.28  & 82.4 & 93.1 & 13.8  & \textbf{3.34}  & \textbf{2.37}  & 3.11  & 4.23  \\
    3D-ViTPose*  & 21.2  & 5.09  & 83.5 & 92.9 & 15.7  & 4.17  & 5.38  & 3.21  & \textbf{4.17}  \\
    LToHP        & \textbf{15.7}  & \textbf{4.12}  & \textbf{89.3} & \textbf{96.1} & \textbf{9.3}   & 3.61  & 3.63  & \textbf{2.82}  & 4.34  \\
    \hline
    \end{tabular}
    }
    \caption{3D posture estimation benchmarks presented in 3D-MuPPET~\cite{3dmuppet}, compared with angular error of the head. 3D estimates are obtained by first detecting keypoints in 2D from multiple views, then triangulated into 3D. Only head keypoints are used to compute error metrics. 3D-KP-RCNN: Keypoint RCNN model~\cite{MaskRCNN}, 3D-DLC*: first detecting pigeons using YOLOv8l\cite{YOLOv8}, then keypoints using DeepLabCut~\cite{DLC}, 3D-ViTPose*: first detecting pigeons using YOLOv8l, then keypoints using ViTPose~\cite{xu2022vitpose}. LToHP: Learnable triangulation of human postures framework~\cite{LToHP}, used as a baseline model for comparison. Bold denotes best model for given metric.}
    \label{tab:method_comparison}
\end{table*}

Next, we show that the model with the best performance in terms of Euclidean distance errors and PCK—LToHP—does not correspond to the best performance when evaluated using head rotation angle, where 3D-DLC* performs better (Table~\ref{tab:method_comparison}). This discrepancy appears to be driven by 3D-DLC*'s lower accuracy in estimating yaw (i.e., horizontal head rotation), which is important if gaze targets are arranged on the same eye-level as the pigeon, which is typically the case.
However, we also find that LToHP yields the lowest RMSE in head rotation angle, suggesting it may produce fewer outliers. In the original 3D-MuPPET publication~\cite{3dmuppet}, the authors recommend 3D-ViTPose* as the most accurate model (excluding LToHP). Yet, based on our evaluation using an application-specific metric, 3D-DLC* appears to perform best. This further highlights the discrepancy between standard ML metrics and application-specific evaluation.


\section{Discussion \& Conclusion}\label{sec:diss}



We presented two case studies that evaluate computer vision models using ML and application-specific metrics. In both cases we report misalignment between metrics, demonstrating that current ML-focused benchmarking is not always optimized to facilitate effective downstream application of models, and complementary mechanisms to evaluate them in this context are needed. Case Study 1 shows that model-derived abundance estimates of chimpanzees diverges notably from those produced using manual annotations, despite high ML performance. Similarly, Case Study 2 shows that the best model for 3D posture estimation is not the best model for estimating head rotation. This highlights the potential for end-users to be misled by impressive ML benchmarks or invest substantial resources (e.g., data collection, annotation, compute~\cite{tuia2022perspectives}) optimising model performance that fails to translate in practice. 



Further analysis of \textit{why} ML and application-specific metrics can be mismatched is needed, though many plausible explanations exist. For example, in Case Study 1 the model is applied in an OOD setting (see Sec.~\ref{sec:setup}) which may degrade performance and ultimately contribute to the mismatch between metrics  -- a key challenge in ML~\cite{beery2018recognition, koh2021wilds, brookes2025fgbg}. 
In case study 2, rotation estimation is likely more sensitive than position estimation, as rotational errors can be compounded in unpredictable ways by small errors in point estimates. These challenges are often found in real-world deployments, and these aspects deserve consideration when developing and benchmarking SOTA algorithms.

Finally, while we present case studies where ML and application-specific metrics can be mismatched, we acknowledge that in certain cases, these metrics can also be aligned. However, this does not downplay the importance of introducing complementary application-specific metrics in current benchmarking pipelines, to better bridge novel algorithms to downstream applications.

\textbf{Future Directions}
To work toward solutions of the presented challenges, we recommend introducing application-specific metrics in existing and new CV datasets with biological/ecological relevance, as complementary benchmarks to traditional ML metrics. While datasets can be used to solve many different problems, researchers could introduce complementary metrics for core problems relating to the proposed application domain of the dataset. This can hopefully encourage novel models to benchmark complementary metrics that better represent the efficacy of a model for downstream applications. Finally, similar to \cite{tuia2022perspectives,pollock2025harnessing}, we encourage more interdisciplinary collaboration between CV researchers and ecologists/biologists, to better align computer vision solutions to real-life problems, which will lead to more holistic solutions as the field continues to grow.

\section*{Acknowledgments}
This work is funded by the Deutsche Forschungsgemeinschaft (DFG, German Research Foundation) under Germany’s Excellence Strategy— EXC 2117—422037984, and DFG project number 15990824. O.B. was supported by the UKRI CDT in Interactive AI (grant EP/S022937/1). U.W. acknowledges funding from the Connected Minds Program, supported by Canada First Research Excellence Fund, Grant \#CFREF-2022-00010.

{\small
\bibliographystyle{ieee_fullname}
\balance
\bibliography{egbib}
}

\end{document}